\RequirePackage{iftex}
\ifXeTeX
  \AddToHook{package/microtype/after}{\renewcommand\DisableLigatures[2][]{}}
\fi
\PassOptionsToPackage{headheight=30pt}{geometry}
\documentclass[10pt,letterpaper]{afu_latex_template/afu}
\usepackage{amsmath,amssymb}
\usepackage[T1]{fontenc}
\usepackage{iftex}
\ifPDFTeX\pdfminorversion=7\fi
\usepackage{array,colortbl}
\usepackage{xspace}
\usepackage[shortlabels,inline]{enumitem}
\usepackage{float,needspace}
\usepackage{bookmark}
\graphicspath{{figures/}{figures/plots/}}


\fancypagestyle{firstpage}{%
  \fancyhf{}%
  \lhead{\includegraphics[height=0.9cm]{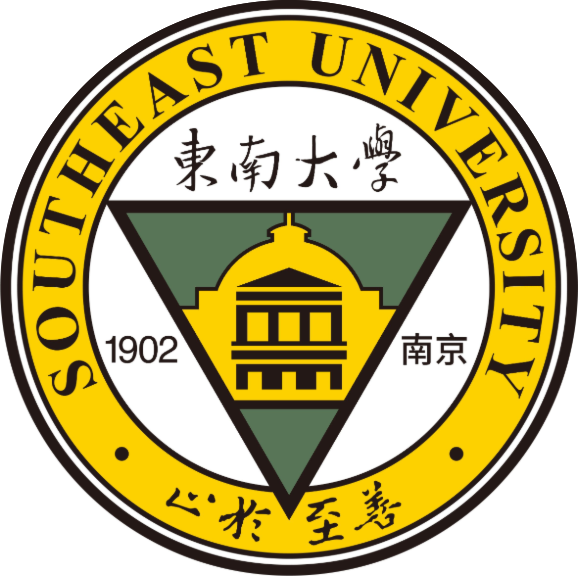}\hspace{0.35cm}\includegraphics[height=0.9cm]{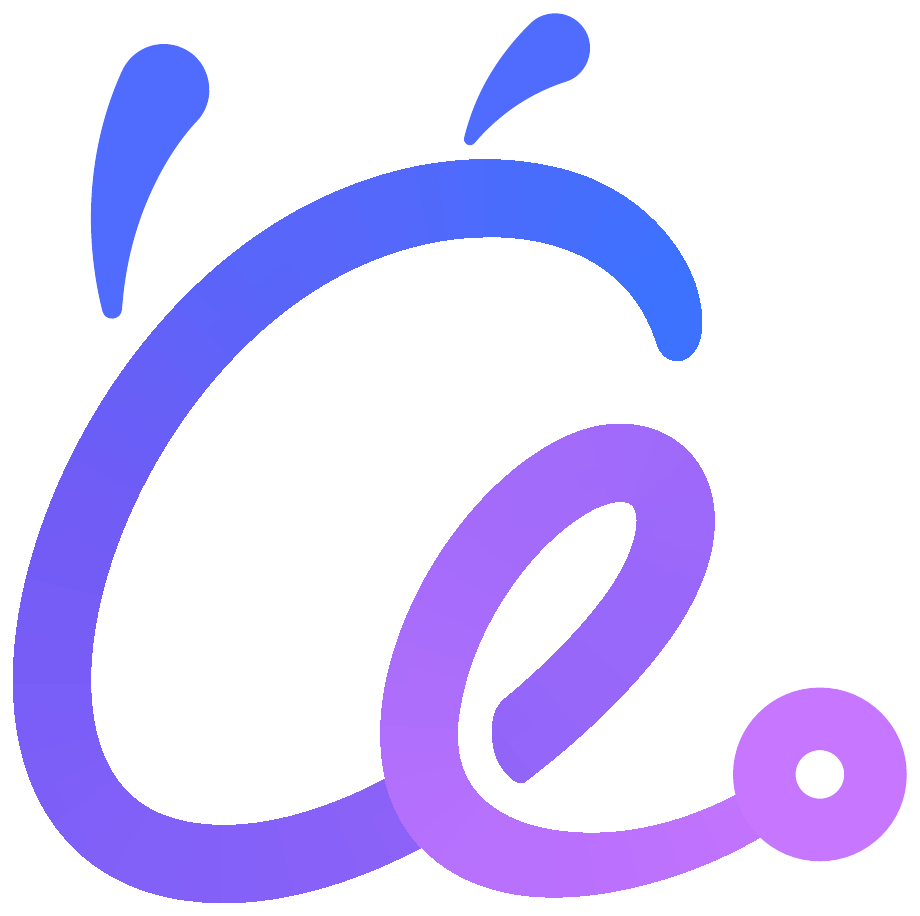}}%
  \rhead{\includegraphics[height=0.9cm]{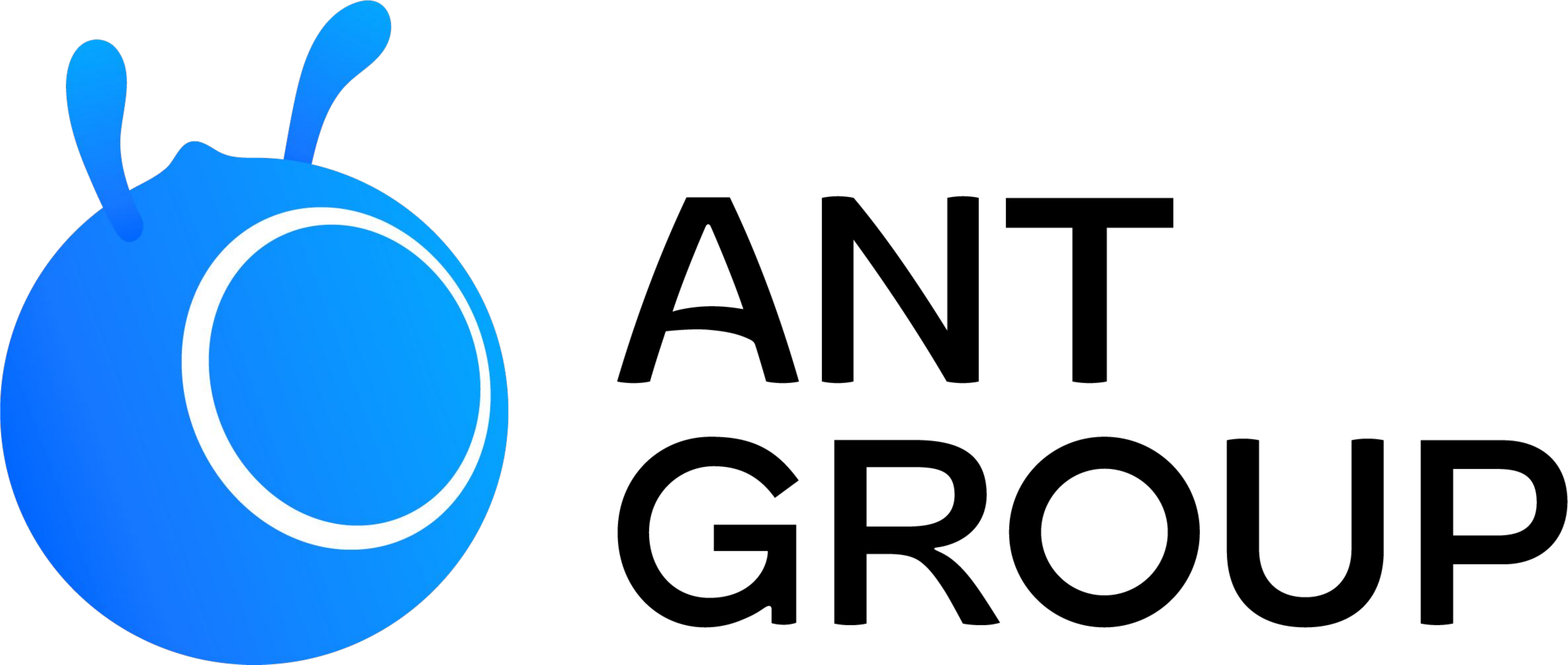}}%
  \fancyfoot[C]{\thepage}%
}

\definecolor{tablegroup}{HTML}{EFF0F2}
\definecolor{tablefocus}{HTML}{EAF2FA}
\definecolor{tablesoft}{HTML}{F6F8FB}
\definecolor{tablehead}{HTML}{F3F5F8}
\makeatletter
\newcommand{\tblrow}[1]{\leavevmode\rlap{\color{#1}\vrule width\linewidth height\ht\@arstrutbox depth\dp\@arstrutbox}}
\makeatother
\definecolor{tablegain}{HTML}{008452}
\definecolor{tableloss}{HTML}{C34427}
\definecolor{tablesep}{HTML}{9CA7B1}
\newcommand{\tdelta}[1]{\,{\normalfont\fontsize{7}{8}\selectfont\ifdim#1pt>0pt\color{tablegain}\else\color{tableloss}\fi(#1)}}
\newcommand{\tsaving}[1]{\,{\normalfont\fontsize{7}{8}\selectfont\color{tablegain}(#1)}}
\newcolumntype{I}{!{\color{tablesep}\vrule width 0.3pt}}
\newcolumntype{P}[1]{>{\raggedright\arraybackslash}p{#1}}
\newcommand{\tgroup}[1]{{\small\bfseries #1}}
\makeatletter
\newcommand{\tblband}{\noalign{\begingroup\color{tablegroup}\hrule height\dimexpr\ht\@arstrutbox+\dp\@arstrutbox\relax\kern-\dimexpr\ht\@arstrutbox+\dp\@arstrutbox\relax\endgroup}}
\makeatother

\title{RubricSearch: Learning Multimodal Agentic Search from Rubric Rewards and Experience-Conditioned Self-Distillation}
\author[1,2,*]{Ming Dai}
\author[2]{Zhihong Lu}
\author[2]{Jinjie Gu}
\author[2]{Jing Hao}
\let\reportauthorfirstrow\authorlist
\renewcommand{\authorlist}{}
\author[1]{Kai Chen}
\author[2]{Jian Wang}
\author[2,\dagger]{Lijun Wu}
\author[1,\dagger]{Wankou Yang}
\let\reportauthorsecondrow\authorlist
\renewcommand{\authorlist}{\reportauthorfirstrow\par\reportauthorsecondrow}
\affiliation[1]{Southeast University}
\affiliation[2]{Ant Group}
\contribution[\dagger]{Corresponding authors}
\correspondence{Lijun Wu (\email{wlj554215@antgroup.com}), Wankou Yang (\email{wkyang@seu.edu.cn})}
\hypersetup{pdftitle={RubricSearch: Learning Multimodal Agentic Search from Rubric Rewards and Experience-Conditioned Self-Distillation},
            pdfauthor={Ming Dai, Zhihong Lu, Jinjie Gu, Jing Hao, Kai Chen, Jian Wang, Lijun Wu, Wankou Yang},
            pdfsubject={Technical Report},
            pdfkeywords={multimodal agentic search, rubric rewards, reinforcement learning, self-distillation},
            bookmarksnumbered=true}
\abstract{
Multimodal agentic search requires an agent to ground visual entities, gather web evidence, and compose a factual answer over many tool calls. Rewarding only final-answer correctness limits such agents in two ways: partial factual progress receives no credit, and group-relative objectives assign one trajectory-level advantage to decisions with very different roles. We present RubricSearch, a training recipe that addresses both limitations. An agent-based pipeline synthesizes visual questions that explicitly request intermediate facts and pairs each fact with a weighted, source-backed rubric criterion. Hierarchical trajectory distillation collects demonstrations by enlarging the search budget only after shorter attempts fail. During online RL, rubric rewards credit partial completion, while experience-conditioned self-distillation refines token-level advantages. A self-teacher sharing the policy's weights reads a same-task action--observation trace and rescores the student's sampled actions and answers; bounded log-probability gaps act as corrections without a learned critic. Experience serves only this auxiliary scoring and is absent from rollout generation and inference. Our analyses show that rubric-supervised progressive tasks raise the fraction of sampled groups with nonzero reward variance from 36.5\% to 57.7\% relative to endpoint-only training. Action--observation traces are better calibration context than reference answers, and moderate corrections applied only to actions and answers work best. Across six public benchmarks, the final 9B policy gains 14.6 percentage points in mean accuracy over tool-enabled Qwen3.5-9B.
}

\begin{document}
\maketitle
% Internship note at the foot of the first page, marked * after the author name.
\begingroup
\renewcommand{\thefootnote}{$*$}
\footnotetext{Work done during Ming Dai's internship at Ant Group.}
\endgroup
\fancyhead[L]{\nouppercase{\textbf{RubricSearch}}}

% !TeX root = ../main.tex
\section{Introduction}
\label{sec:introduction}

Multimodal agentic search combines visual understanding with open-world evidence gathering: agents recognize image entities, issue text and image queries, read webpages, and assemble retrieved facts into answers. They acquire this behavior through prompted tool use, trajectory imitation, and interactive reinforcement learning (RL)~\citep{yao2022react,nakano2022webgptbrowserassistedquestionansweringhuman,jin2025searchr1trainingllmsreason,song2025r1searcherincentivizingsearchcapability,zheng2025deepresearcherscalingdeepresearch,li2025webthinkerempoweringlargereasoning,du2026openseeker}. Recent multimodal systems coordinate visual grounding, search, and webpage interaction through these recipes~\citep{mmsearch-r1,webwatcher,deepeyesv2,chng2025sensenova,huang2026vision,chen2026opensearch,jiao2026searcheyes,wei2026unibrowse}. Their RL rewards mainly reflect final-answer correctness, creating three limitations.

\begin{figure}[t]
  \centering
  \includegraphics[width=\linewidth]{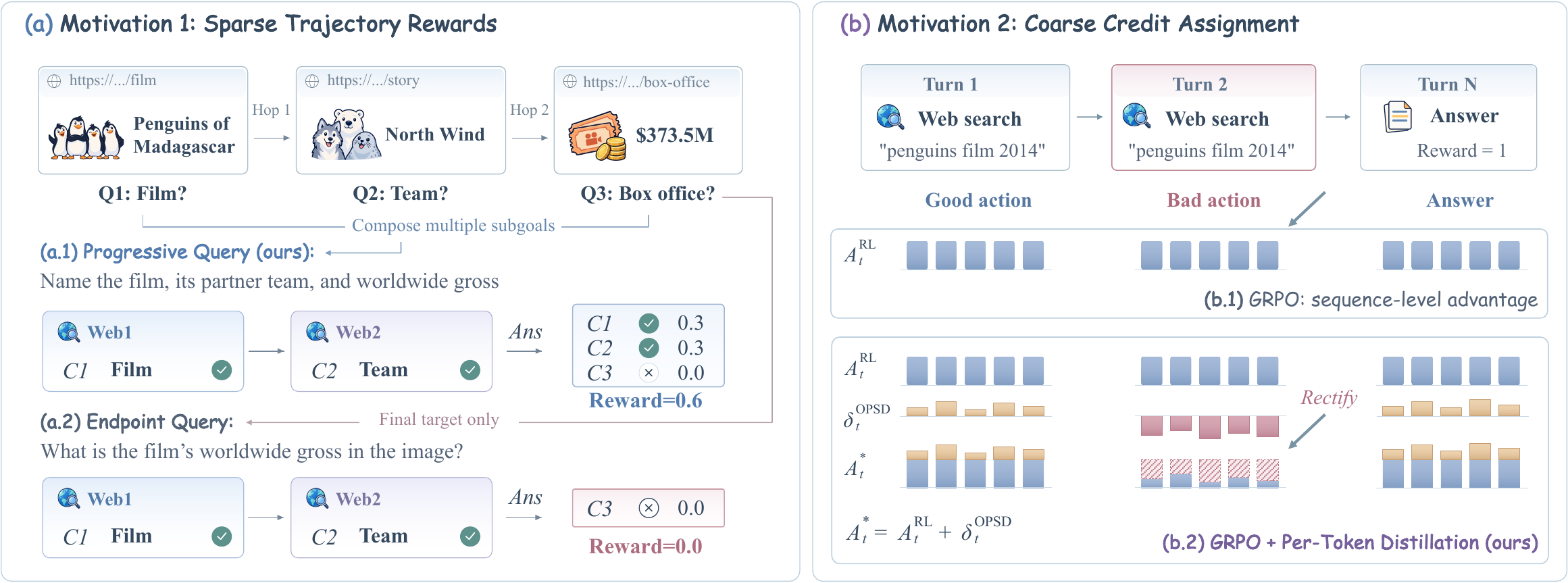}
  \caption{\textbf{Two feedback bottlenecks.} (a) Progressive factual targets expose partial completion through weighted rubrics. (b) Experience-conditioned corrections distinguish actions and answers that otherwise receive the same trajectory-level RL advantage.}
  \label{fig:motivation}
\end{figure}

\Needspace{4\baselineskip}
\textbf{(i) Sparse outcome rewards.} In Fig.~\ref{fig:motivation}(a), an agent identifies a film and its partner team but misses the worldwide gross; an endpoint-only reward scores this answer like one that recovers nothing. Several search RL recipes use terminal answer-quality rewards, sometimes with format rewards~\citep{jin2025searchr1trainingllmsreason,webwatcher}. When all sampled answers are equally wrong, group-relative training gets no signal. Process supervision~\citep{deng2025atomsearcher,wen2026smartsearch} instead requires additional per-step judgments.

\Needspace{4\baselineskip}
\textbf{(ii) Coarse credit assignment.} A correct answer may follow redundant searches, while a well-chosen query can still end in failure. Outcome-supervised GRPO~\citep{shao2024deepseekmath} and related critic-free objectives~\citep{yu2025dapo,liu2025understandingr1zero} assign one trajectory-level advantage to every generated token, so the repeated search in Fig.~\ref{fig:motivation}(b) inherits the credit of the final answer. Actor--critic methods based on PPO~\citep{schulman2017ppo}, including VAPO~\citep{yue2025vapo} and SAO~\citep{hou2026sao}, obtain finer feedback from learned values and GAE~\citep{schulman2015gae}, but must train a critic over long multimodal contexts. On-policy self-distillation (OPSD)~\citep{zhao2026selfdistilledreasoneronpolicyselfdistillation} and feedback-conditioned variants~\citep{hueb2026reinforcementlearningselfdistillation,wang2026skillsdskillconditionedselfdistillationmultiturn,lu2026selfdistilledagenticreinforcementlearning} avoid the critic: the same model, given richer training-time context, scores the student's sampled tokens. Recent work studies critic alignment~\citep{fu2026pact} and feedback selection~\citep{li2026distillselectivehindsightdistillation,wang2026risedrubricsagenticmultienvironment,zhou2026selfevolvingagentsprivilegedretrospective}, but the choice of experience context for assigning credit to search actions remains open.

\Needspace{4\baselineskip}
\textbf{(iii) Unregulated interaction cost.} WebThinker~\citep{li2025webthinkerempoweringlargereasoning}, LMM-Searcher~\citep{du2026towards}, and DeepVoyager-VL~\citep{zhang2026deepvoyagervlincentivizingvisionintheloopsearch} advance long-horizon search, yet interactive uses such as product lookup and customer support are sensitive to latency and cost. Outcome rewards ignore how many calls a correct answer consumes, so agents may spend extra search rounds and tool operations without improving accuracy.

\begin{figure}[t]
  \centering
  \includegraphics[width=\textwidth]{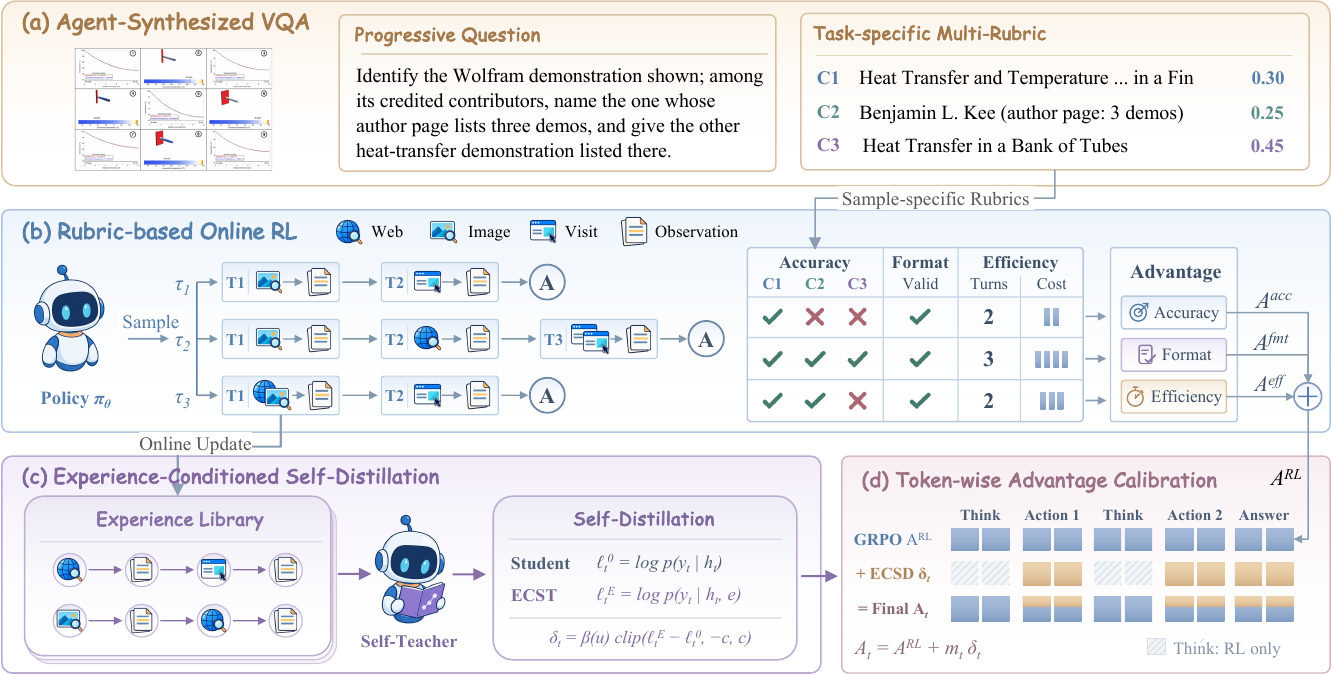}
  \caption{\textbf{RubricSearch overview.} Online RL uses synthesized tasks and rubrics. A self-teacher uses experience for bounded action--answer corrections; rollouts and deployment receive no experience.}
  \label{fig:framework}
\end{figure}

RubricSearch addresses these limitations (Fig.~\ref{fig:framework}). \textbf{(i)}~An agent-based pipeline synthesizes progressive visual questions and attaches a weighted, source-backed criterion to every requested fact, so rewards credit partial completion from the answer alone. Unlike instance-specific rubrics written for existing prompts~\citep{gunjal2025rubricsrewardsreinforcementlearning,viswanathan2025checklistsbetterrewardmodels,shao2025drtulu,zhang-etal-2026-chaining}, questions and criteria are generated jointly. \textbf{(ii)}~In experience-conditioned self-distillation, a self-teacher sharing the student's weights reads a same-task action--observation trace and rescores sampled actions and answers; bounded log-probability gaps refine their RL advantages without a critic. Rollouts and deployment never see experience. \textbf{(iii)}~Effort enters at three points: hierarchical distillation keeps the first success under increasing budgets, calibration uses the lowest-effort successful reference, and optional bonuses prefer cheaper correct solutions. A cascade keeps judging affordable: Jev~\citep{almeida2026jev} scores first; an LLM rejudges uncertain cases.

Our analyses yield three findings. First, rubric-supervised tasks give nonzero group-relative feedback for many more sampled groups. Second, action--observation traces are better calibration context than reference answers, which help as prompts but hurt as training feedback. Third, corrections work best at moderate strength and when restricted to actions and answers. Our contributions are as follows:\vspace{-3pt}
\begin{itemize}[leftmargin=*,topsep=0pt,itemsep=1pt,parsep=0pt]
  \item \textbf{Rubric-supervised task synthesis.} We jointly synthesize progressive visual questions with weighted, source-backed rubrics and hierarchically distill compact SFT demonstrations.
  \item \textbf{Experience-conditioned self-distillation.} Same-task traces yield bounded token-level corrections to the RL advantages of individual sampled tool actions and answers.
  \item \textbf{Feedback analysis.} We study reward sparsity, experience content, calibration scope and strength, resource bonuses, and the cost of rubric judging during training.
  \item \textbf{Empirical results.} On six benchmarks, RubricSearch-9B gains 14.6 points in mean accuracy over tool-enabled Qwen3.5-9B~\citep{qwen2026qwen35}; both scales are competitive with recent multimodal agents.
\end{itemize}

% !TeX root = ../main.tex
\section{Related Work}
\label{sec:related}

\paragraph{Multimodal agentic search.}
Multimodal agents learn visual grounding, retrieval, and webpage interaction from supervised trajectories and online RL~\citep{mmsearch-r1,webwatcher,deepeyesv2,chng2025sensenova,huang2026vision,zhang2026visualseeker}. Training pipelines verify evidence and simulate search environments~\citep{chen2026opensearch,dai2026simplesearchvlsimplerecipemultimodal,wei2026unibrowse,jiao2026searcheyes}. DeepVoyager-VL~\citep{zhang2026deepvoyagervlincentivizingvisionintheloopsearch} and Neos~\citep{he2026neos} train agents to use intermediate visual evidence for continued search; the latter uses successful references to assign credit. We combine rubric rewards and bounded token-level action--answer calibration.

\paragraph{Rubric-based supervision.}
Criterion-based evaluators assess responses along explicit quality dimensions~\citep{liu-etal-2023-g,kim2024prometheusinducingfinegrainedevaluation}. Instruction-specific checklists and rubric rewards make these criteria usable for open-ended policy optimization~\citep{gunjal2025rubricsrewardsreinforcementlearning,viswanathan2025checklistsbetterrewardmodels,huang2025reinforcementlearningrubricanchors}. For deep search, DR Tulu~\citep{shao2025drtulu} and CaRR~\citep{zhang-etal-2026-chaining} use search evidence to reward coverage and grounding. Retrieved knowledge supports rubric construction~\citep{mei2026deepresearchrubricreinforcement}, and expert criteria evaluate research reports~\citep{sharma2025researchrubricsbenchmarkpromptsrubrics}. Multimodal extensions supervise perception, grounding, and reasoning consistency~\citep{chen2026ruclstratifiedrubricbasedcurriculum,tian2026vrubricsvisualfaithfulnessrubricbased}. RubricSearch jointly constructs image-dependent questions and source-backed criteria, tying each criterion to an explicitly requested fact.

\paragraph{On-policy self-distillation.}
Distillation on student-generated sequences reduces the mismatch between fixed demonstrations and the learner's sampling distribution~\citep{gu2024minillm,agarwal2023gkd}. Self-distillation obtains token-level feedback from the same model under richer training-time context~\citep{zhao2026selfdistilledreasoneronpolicyselfdistillation,hueb2026reinforcementlearningselfdistillation}. Rubrics, reusable skills, and retrospective lessons supply structured context for open-ended and multi-turn tasks~\citep{bablani2026rubricsprivilegedinformationopenended,wang2026skillsdskillconditionedselfdistillationmultiturn,wang2026risedrubricsagenticmultienvironment,zhou2026selfevolvingagentsprivilegedretrospective}. For multimodal search, MMSearch-OPD~\citep{zhang2026mmsearchopdeasyhardhard} combines RL with reward-gap-modulated token-level teacher guidance. SERL~\citep{li2026distillselectivehindsightdistillation} integrates hindsight preferences through bounded, sign-preserving reweighting of reward-derived action advantages. RubricSearch adds bounded action--answer corrections derived from same-task successes or rubric-annotated failures to rubric-based RL, and studies which experience content and which token types benefit from such calibration; rollout generation and inference never receive experience as input.

% !TeX root = ../main.tex
\section{Method}
\label{sec:method}

RubricSearch learns from factual rewards and experience-conditioned calibration, with a shared data pipeline (Fig.~\ref{fig:framework}). For visual question $x=(q,\mathcal I)$, policy $\pi_\theta$ produces trajectory $\tau_i$ and answer $a_i$ through reasoning, actions, and observations. Each task has a reference $a^\star$ and rubric $\mathcal R=\{(c_k,w_k)\}_{k=1}^{K}$, where $c_k$ is a factual target and $w_k>0$ its weight, with $\sum_k w_k=1$. The rollout policy receives the question and images, without reference annotations. Token $y_{i,t}$ is generated after prefix $h_{i,t}$.

\begin{figure}[t]
  \centering
  \includegraphics[width=\textwidth]{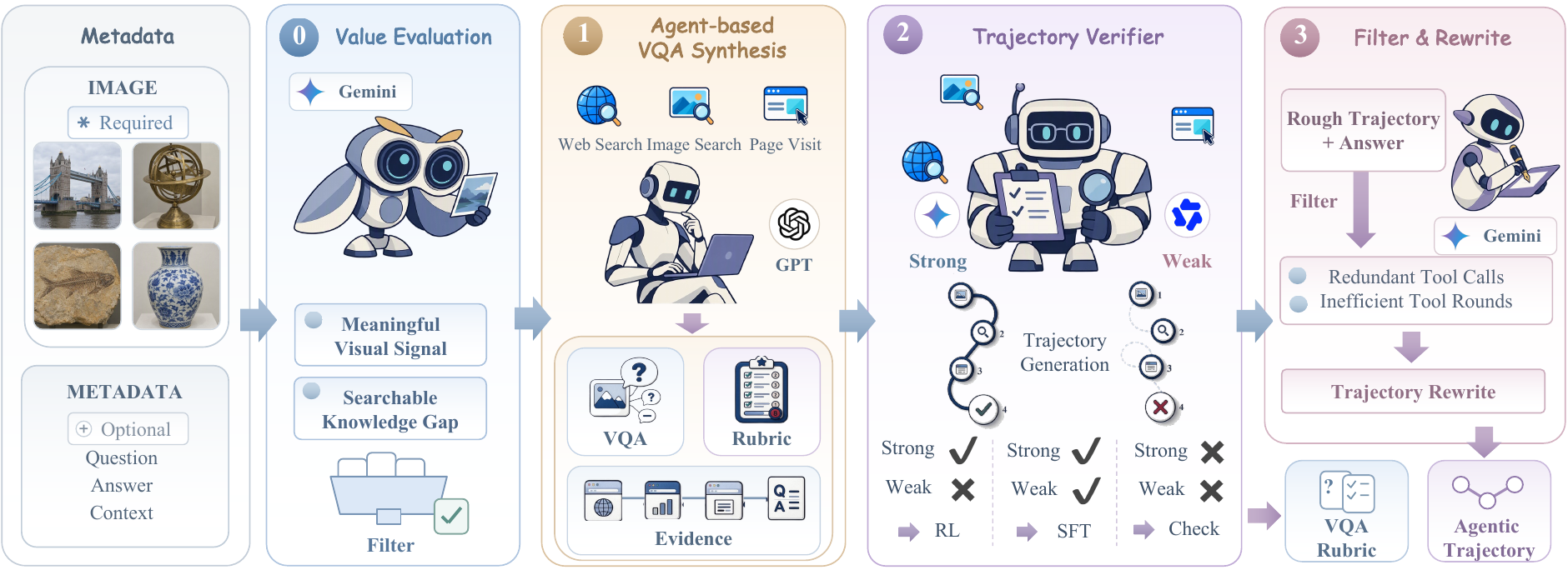}
  \caption{\textbf{Training-data synthesis.} Evidence gathering produces visual questions, answers, and weighted criteria. Solver verification routes tasks to SFT, RL, or inspection; curation preserves executed actions and the evidence available at each step.}
  \label{fig:data-synthesis}
\end{figure}

\subsection{Agent-Based Data Synthesis}
\label{sec:data-synthesis}
The pipeline defines verifiable rubric targets and collects compact demonstrations for supervised initialization. Synthesis, trajectory-generation, and verification agents coordinate four stages (Fig.~\ref{fig:data-synthesis}). Step~0 deduplicates images and screens for search-relevant cues. Step~1 explores image--metadata pairs to construct source-backed questions, answers, and weighted rubrics. Step~2 verifies tasks with stronger and weaker solvers under increasing turn budgets, collecting demonstrations and routing examples to SFT or RL. Step~3 curates reasoning without changing actions or observations.

\subsubsection{Agent-Based VQA Synthesis}
Graph-based synthesis builds questions around entity paths: OpenSearch-VL~\citep{chen2026opensearch} samples Wikipedia hyperlinks, SearchEyes~\citep{jiao2026searcheyes} uses typed knowledge-graph paths, and UniBrowse~\citep{wei2026unibrowse} adds live retrieval. To establish source-backed targets before fixing questions, a synthesis agent based on GPT-5.5~\citep{openai2026gpt55} explores each image's metadata-related context using reverse image search, text search, and webpage visits. After full-image search, research-only prompts broaden exploration across entities, relationships, chronology, and works. Metadata only suggests leads; every target must be substantiated by a visited source. The agent selects an evidence chain and writes an image-dependent, multi-hop question, reference answer, and one weighted criterion per requested fact. Intermediate facts are asked for rather than given as premises, so partial progress is measurable while the search route stays open.

\subsubsection{Hierarchical Trajectory Distillation}
\label{sec:hierarchical-distillation}
Trajectory collection starts with a small search budget and expands it only after failure. After a no-tool check separates directly answerable examples, stronger and weaker agents attempt the remaining questions with 2, 4, 6, 8, and 10 assistant turns. Each solver retains its first correct, format-valid trace and skips larger budgets once it succeeds. Questions solved by both yield SFT demonstrations; those solved by only one enter RL with answers and rubrics. Joint failures require inspection. Curation rejects unsupported or repetitive traces and refines reasoning without changing the action--observation sequence.

\subsection{Rubric-Based Online RL}
\label{sec:rubric-rl}

Online RL rewards factual coverage and valid interaction. An optional advantage bonus compares resource use among fully correct, format-valid solutions within each question's sampled group.

\subsubsection{Rubric and Format Rewards}
\label{sec:rubric-format-rewards}
Following rubric reward learning~\citep{gunjal2025rubricsrewardsreinforcementlearning,shao2025drtulu}, we score requested facts separately to provide GRPO~\citep{shao2024deepseekmath} with graded feedback. Normalized exact matches receive full credit without model-based judging. Otherwise, Jev~\citep{almeida2026jev} predicts criterion-satisfaction probabilities $p_{i,k}$ from the question, reference, candidate answer, and rubric. If any probability lies in $[0.3,0.7]$, an LLM rejudges all criteria and replaces Jev's decisions; otherwise $z_{i,k}=\mathbf 1[p_{i,k}\geq0.5]$. Rubric credit and complete correctness are
\begin{equation}
 s_i=\sum_{k=1}^{K} w_k z_{i,k},\qquad
 v_i=\prod_{k=1}^{K} z_{i,k}.
 \label{eq:rubric-credit}
\end{equation}
Here $s_i$ measures partial completion and $v_i$ complete correctness; neither labels actions. For $G$ trajectories per question, $r_i=\alpha_s s_i+\alpha_f f_i$ combines factual credit and format validity $f_i$ with positive weights $\alpha_s,\alpha_f$. The latter rewards valid interaction and answer termination; unusable answers receive zero. Group normalization gives outcome advantage $A_i^{\mathrm{out}}$.

\subsubsection{Success-Conditioned Resource Bonuses}
\label{sec:efficiency-advantages}
Accuracy-only rewards are indifferent to how much interaction a correct answer consumes. To prefer economical success without rewarding cheap failure, resource use $d_i$ is compared only among fully correct, format-valid trajectories $\mathcal S$ within each question's sampled group:
\begin{equation}
 b_i(d)=\frac{\bar d_{\mathcal S}-d_i}
 {d_{\mathcal S}^{\max}-d_{\mathcal S}^{\min}},\quad i\in\mathcal S,
 \label{eq:efficiency-advantage}
\end{equation}
Here $\bar d_{\mathcal S}$ is the successful-group mean; the bonus is zero outside $\mathcal S$, for $|\mathcal S|<2$, or for zero range. We count assistant turns $T_i$ and logical tool cost $C_i$, a weighted count of requested text queries, image regions, and distinct webpage targets within calls. These nonmonetary resources give
\begin{equation}
 A_i^{\mathrm{RL}}=A_i^{\mathrm{out}}+
 \lambda_T b_i(T)+\lambda_C b_i(C).
 \label{eq:rl-advantage}
\end{equation}
Nonnegative $\lambda_T,\lambda_C$ weight turn and tool-cost bonuses, respectively. These bounded terms average to zero over $\mathcal S$, leave rewards unchanged, and are disabled in most analyses of Sec.~\ref{sec:analysis-feedback}.

\subsection{Experience-Conditioned Self-Distillation}
\label{sec:distillation}

Rubric rewards and resource bonuses distinguish trajectories but assign a shared advantage to their tokens. Same-task experience refines this shared signal on actions and answers.

\subsubsection{Privileged Information Construction}
\label{sec:privileged-information}
The task-indexed experience library stores validated attempts, including those from the current batch, while excluding execution and judging errors. Among successful entries, it selects the reference $\tau^\star$ with minimum effort $J(\tau)=\eta_T T(\tau)+\eta_C C(\tau)$; nonnegative $\eta_T,\eta_C$ weight turns and logical tool cost for reference selection. Failed attempts can use it directly, whereas successful attempts must satisfy
\begin{equation}
 J(\tau_i)-J(\tau^\star)>
 \max\{\kappa_a,\kappa_r J(\tau^\star)\},
 \label{eq:experience-eligibility}
\end{equation}
where $\kappa_a,\kappa_r\geq0$ set absolute and relative cost margins. If no validated success is available, the library selects the highest-credit failure, breaking ties by greater effort.

Privileged information (PI) $e_i$ contains ordered actions and compact observations, with reasoning and final-answer text removed. Observations can still contain evidence for the answer. Failed references are explicitly marked and list unmet criteria, so the scorer can distinguish useful evidence from an unsuccessful outcome. Only auxiliary scoring uses PI; over-budget references are skipped.

\subsubsection{Token-Level Advantage Calibration}
\label{sec:token-advantage-calibration}
Following feedback-conditioned self-distillation~\citep{hueb2026reinforcementlearningselfdistillation,zhao2026selfdistilledreasoneronpolicyselfdistillation}, RubricSearch refines credit on sampled trajectories. Its experience-conditioned self-teacher shares the student's weights and is refreshed after every policy update. Let $\pi_{\bar\theta}$ denote its synchronized snapshot and $\pi_{\mathrm{old}}$ the pre-update student policy. Both score the same sampled tokens at their original prefixes, but only the teacher receives PI:
\begin{equation}
 \begin{aligned}
 \ell^E_{i,t}&=\log\pi_{\bar\theta}(y_{i,t}\mid h_{i,t},e_i),\\
 \ell^0_{i,t}&=\log\pi_{\mathrm{old}}(y_{i,t}\mid h_{i,t}),\\
 \delta_{i,t}&=\beta(u)\,\mathrm{clip}
 (\ell^E_{i,t}-\ell^0_{i,t},-c,c).
 \end{aligned}
 \label{eq:experience-correction}
\end{equation}
Here $c>0$ bounds the log-probability gap. Guidance $\beta(u)=\beta_{\max}\min\{1,(u+1)/W\}$ increases linearly to $\beta_{\max}$ over $W$ updates, with zero-based update index $u$. Missing or ineligible experience gives $\delta_{i,t}=0$. The gap measures how experience changes token likelihood, not action value.

Binary mask $m_{i,t}$ selects only tool-action tokens (including arguments) and final-answer tokens. Reasoning tokens keep the original RL advantage, and tool observations are excluded from the loss. The resulting advantage adds this masked correction to the base RL signal:
\begin{equation}
 A^{\mathrm{final}}_{i,t}=A_i^{\mathrm{RL}}
 +m_{i,t}\,\operatorname{stopgrad}(\delta_{i,t}).
 \label{eq:advantage-fusion}
\end{equation}
Stop-gradient holds $\delta_{i,t}$ constant during optimization. With likelihood ratio $\rho_{i,t}=\pi_\theta(y_{i,t}\mid h_{i,t})/\pi_{\mathrm{old}}(y_{i,t}\mid h_{i,t})$, the clipped surrogate trains on sampled reasoning, action, and answer tokens:
\begin{equation}
 \begin{aligned}
 \mathcal L(\theta)&=-\mathbb E\Big[\min\big(\rho_{i,t}A^{\mathrm{final}}_{i,t}, \bar\rho_{i,t}A^{\mathrm{final}}_{i,t}\big)\Big],\\
 \bar\rho_{i,t}&=\mathrm{clip}(\rho_{i,t},1-\epsilon_l,1+\epsilon_h).
 \end{aligned}
 \label{eq:policy-objective}
\end{equation}
Here $\epsilon_l,\epsilon_h$ are lower and upper clipping widths. Corrections satisfy $|\delta_{i,t}|\leq\beta(u)c$, and $\beta=0$ recovers rubric-based RL. Rollouts and deployment receive no PI, and inference uses no auxiliary LLM.

% !TeX root = ../main.tex
% Single-column report; floats are placed at the top of pages.

\section{Experiments}
\label{sec:experiments}

\subsection{Experimental Setup}
\label{sec:experimental-setup}
RubricSearch uses the 9B and 35B-A3B variants of Qwen3.5~\citep{qwen2026qwen35} as policy backbones. SFT uses 7,479 curated demonstrations; RL uses 2,161 synthesized multi-rubric tasks. Slime~\citep{slime_github} trains with a 64K-token context; each update samples four trajectories per question, for 64 questions in main runs and 32 in analysis runs. Asynchronous training combines partial rollouts and optimizer-state CPU offloading on one server with eight NVIDIA H200 GPUs at either scale. Analysis runs use 60 updates; main runs use 150 updates with resource-bonus weights $\lambda_T=\lambda_C=0.2$; the 9B model completes 150 RL updates in under 20 hours. Training and evaluation share a ten-turn limit.

Main experiments evaluate the complete test sets of MMSearch~\citep{jiang2024mmsearchbenchmarkingpotentiallarge} (image-only), MMSearch+~\citep{tao2025mmsearch}, BrowseComp-VL~\citep{webwatcher}, FVQA~\citep{wang2017fvqa}, LiveVQA~\citep{fu2025seekingupdatinglivevisual}, and SimpleVQA~\citep{cheng2025simplevqa}. Ablations retain all 171 MMSearch Image questions and use 100-question subsets of BrowseComp-VL, MMSearch+, FVQA, and LiveVQA; the first three form the 371-question search subset. Gemini 3.5 Flash judges answers against benchmark references, with failures retained in accuracy denominators. Turns count assistant turns per trajectory, and logical tool cost counts text queries, image regions, and visited URLs. At evaluation, only the frozen-SFT diagnostic receives experience.

\subsection{Main Results}
\label{sec:main-results}

\begin{table}[t]
    \centering
    \caption{\textbf{Accuracy on six multimodal search benchmarks (\%).} MMSearch denotes its image split; $\ddagger$ marks results on the full MMSearch set with text and image queries combined. $\dagger$: evaluated on a 300-question subset in the source paper. \textsuperscript{\S}: no auxiliary LLM at inference; ``--'': not reported; bold: best agent per block; parentheses: gain over the preceding Base or SFT row (pp).}
    \label{tab:search-qa-main}
    \footnotesize
    \setlength{\tabcolsep}{2pt}
    \renewcommand{\arraystretch}{1.14}
    \begin{tabular*}{\textwidth}{@{}@{\extracolsep{\fill}}llIcccccc@{}}
        \toprule
        \textbf{Method} & \textbf{Base model} & \textbf{MMSearch} & \textbf{MMSearch+} & \textbf{BC-VL} & \textbf{FVQA} & \textbf{LiveVQA} & \textbf{SimpleVQA} \\
        \midrule
        \rowcolor{tablegroup}\multicolumn{8}{c}{\tgroup{Direct Answer}} \\
        GPT-5.6-Sol~\citep{openai2026gpt56sol} & - & 68.4 & 21.5 & 47.4 & 63.1 & 72.9 & 72.3 \\
        Gemini-3.1-Pro~\citep{google2026gemini31pro} & - & 60.8 & 24.1 & 42.1 & 56.7 & 63.3 & 68.7 \\
        Kimi-K3~\citep{moonshot2026kimik3} & - & 65.5 & 17.0 & 40.9 & 60.3 & 72.3 & 70.0 \\
        Qwen3.5~\citep{qwen2026qwen35} & Qwen3.5-9B & 29.2 & 2.4 & 25.1 & 28.0 & 50.6 & 54.6 \\
        Qwen3.5~\citep{qwen2026qwen35} & Qwen3.5-35B-A3B & 32.2 & 4.5 & 30.6 & 34.9 & 56.3 & 58.2 \\
        \midrule
        \rowcolor{tablegroup}\multicolumn{8}{c}{\tgroup{Multimodal Search Agent}} \\
        SenseNova-MARS~\citep{chng2025sensenova} & Qwen3-VL-8B & 67.8 & -- & -- & 67.1 & 56.2 & 70.2 \\
        Vision-DeepResearch~\citep{huang2026vision} & Qwen3-VL-8B & 69.6 & 20.4 & 42.6 & 64.7$^{\dagger}$ & 76.7$^{\dagger}$ & -- \\
        OpenSearch-VL~\citep{chen2026opensearch} & Qwen3-VL-8B & 64.5 & -- & 37.6 & 71.5 & 59.6 & 71.6 \\
        SimpleSearch-VL~\citep{dai2026simplesearchvlsimplerecipemultimodal}\textsuperscript{\S} & Qwen3-VL-8B & 77.1 & 32.5 & 52.1 & 76.8 & 75.2 & 76.6 \\
        SearchEyes~\citep{jiao2026searcheyes} & Qwen3.5-9B & 69.2 & -- & 42.3 & 74.2 & 66.1 & 75.4 \\
        Pistis-Agentic~\citep{chen2026pististechnicalreport} & Qwen3.5-9B & 72.7$^{\ddagger}$ & -- & 53.2 & -- & \textbf{82.7} & -- \\
        Qwen3.5~\citep{qwen2026qwen35} (Base) & Qwen3.5-9B & 65.5 & 26.0 & 44.4 & 67.3 & 60.6 & 64.3 \\
        \tblrow{tablesoft}\textbf{RubricSearch} (SFT)\textsuperscript{\S} & Qwen3.5-9B & 78.4\tdelta{+12.9} & 33.8\tdelta{+7.8} & 51.4\tdelta{+7.0} & 75.2\tdelta{+7.9} & 68.7\tdelta{+8.1} & 75.3\tdelta{+11.0} \\
        \tblrow{tablefocus}\textbf{RubricSearch}\textsuperscript{\S} & Qwen3.5-9B & \textbf{81.9}\tdelta{+3.5} & \textbf{38.3}\tdelta{+4.5} & \textbf{57.9}\tdelta{+6.5} & \textbf{80.0}\tdelta{+4.8} & 78.0\tdelta{+9.3} & \textbf{79.4}\tdelta{+4.1} \\
        \midrule
        Vision-DeepResearch~\citep{huang2026vision} & Qwen3-VL-30B-A3B & 69.6 & 28.5 & 53.7 & 74.2$^{\dagger}$ & 77.6$^{\dagger}$ & -- \\
        OpenSearch-VL~\citep{chen2026opensearch} & Qwen3-VL-30B-A3B & 68.7 & -- & 41.1 & 73.2 & 67.4 & 74.9 \\
        SimpleSearch-VL~\citep{dai2026simplesearchvlsimplerecipemultimodal}\textsuperscript{\S} & Qwen3-VL-30B-A3B & \textbf{83.6} & 34.4 & 55.9 & 79.0 & 81.1 & 79.6 \\
        WeAgent-MMSearch~\citep{liu2026weagentmmsearchnativetextvisioninteraction} & Qwen3-VL-30B-A3B & 77.8 & 37.8 & -- & 79.3 & 81.1 & 77.7 \\
        UniBrowse~\citep{wei2026unibrowse} & Qwen3.5-35B-A3B & 75.2 & 34.1 & 57.8 & 76.7 & -- & -- \\
        SearchEyes~\citep{jiao2026searcheyes} & Qwen3.5-27B & 82.4 & -- & 49.3 & 79.1 & 77.3 & 80.9 \\
        Qwen3.5~\citep{qwen2026qwen35} (Base) & Qwen3.5-35B-A3B & 75.4 & 32.8 & 48.1 & 70.3 & 64.2 & 69.2 \\
        \tblrow{tablesoft}\textbf{RubricSearch} (SFT)\textsuperscript{\S} & Qwen3.5-35B-A3B & 77.8\tdelta{+2.4} & 37.3\tdelta{+4.5} & 56.4\tdelta{+8.3} & 77.1\tdelta{+6.8} & 72.9\tdelta{+8.7} & 77.1\tdelta{+7.9} \\
        \tblrow{tablefocus}\textbf{RubricSearch}\textsuperscript{\S} & Qwen3.5-35B-A3B & \textbf{83.6}\tdelta{+5.8} & \textbf{41.8}\tdelta{+4.5} & \textbf{61.4}\tdelta{+5.0} & \textbf{82.1}\tdelta{+5.0} & \textbf{83.1}\tdelta{+10.2} & \textbf{82.0}\tdelta{+4.9} \\
        \bottomrule
    \end{tabular*}
\end{table}

Table~\ref{tab:search-qa-main} groups our Base, SFT, and final policies at each scale. Averaged equally over six benchmarks, the final 9B policy gains 14.6 percentage points over the tool-enabled Qwen3.5-9B base and 5.5 points over its SFT initialization. Online training thus improves on tool access and supervised imitation. The 9B policy leads its block on five benchmarks, trailing Pistis-Agentic~\citep{chen2026pististechnicalreport} on LiveVQA. The 35B policy ties SimpleSearch-VL~\citep{dai2026simplesearchvlsimplerecipemultimodal} on MMSearch Image and leads on the other five benchmarks.

We compare systems under their own protocols. SenseNova-MARS~\citep{chng2025sensenova} and Vision-DeepResearch~\citep{huang2026vision} summarize retrieved content with larger auxiliary models. Like SimpleSearch-VL~\citep{dai2026simplesearchvlsimplerecipemultimodal}, RubricSearch uses the policy itself to summarize visited webpages, so inference requires no auxiliary language model while remaining competitive across all six evaluated benchmarks.

\subsection{What Makes the Feedback Effective?}
\label{sec:ablation}
\label{sec:analysis-feedback}

We examine the two signals through targeted variations of the training recipe. Unless stated otherwise, analysis runs train the 9B policy with $\beta=0.1$, action--answer corrections, and no resource bonus; AVG pools the 371 search questions with weights set by subset size.

\begin{figure}[!b]
    \centering
    \begin{subfigure}[t]{0.31\textwidth}
        \centering
        \begin{minipage}[t][82bp][t]{\linewidth}
            \vspace{3.5bp}
            \centering
            \scriptsize
            \renewcommand{\arraystretch}{1.26}
            \setlength{\tabcolsep}{1.2pt}
            \begin{tabular*}{\linewidth}{@{}@{\extracolsep{\fill}}lcccc@{}}
                \toprule
                \textbf{Method} & \textbf{AVG} & \textbf{MMS} & \textbf{BC-VL} & \textbf{MMS+} \\
                \midrule
                RL ($\beta=0$) & 59.8 & 78.9 & 56.0 & 31.0 \\
                $\beta=0.05$ & 62.5 & \textbf{80.1} & 61.0 & 34.0 \\
                \tblrow{tablefocus}$\beta=0.1$ & \textbf{62.8} & 79.5 & \textbf{62.0} & \textbf{35.0} \\
                $\beta=0.2$ & 58.5 & 77.8 & 59.0 & 25.0 \\
                \bottomrule
            \end{tabular*}
        \end{minipage}
        \caption{Search accuracy (\%).}
        \label{fig:opsd-beta-metrics}
    \end{subfigure}\hspace{0.008\textwidth}%
    \begin{subfigure}[t]{0.315\textwidth}
        \centering
        \begin{minipage}[t][82bp][t]{\linewidth}
            \vspace{0pt}
            \centering
            \includegraphics[width=\linewidth]{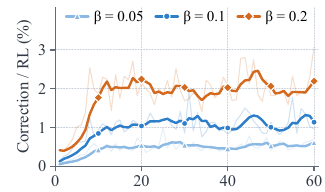}
        \end{minipage}
        \caption{Advantage correction.}
        \label{fig:opsd-beta-strength}
    \end{subfigure}\hspace{0.008\textwidth}%
    \begin{subfigure}[t]{0.359\textwidth}
        \centering
        \begin{minipage}[t][82bp][t]{\linewidth}
            \vspace{0pt}
            \centering
            \includegraphics[width=\linewidth]{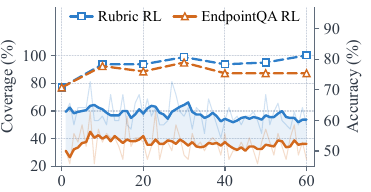}
        \end{minipage}
        \caption{Rubric supervision.}
        \label{fig:multirubric-coverage}
    \end{subfigure}
    \caption{\textbf{Guidance strength and rubric supervision.} Thin/thick lines show per-update values/five-update means; (b) uses identical tokens. In (c), solid coverage and dashed MMSearch Image accuracy use independent scales; shading shows the coverage gap between rubric and endpoint-only RL.}
    \label{fig:opsd-beta}
\end{figure}

\subsubsection{Does Rubric Feedback Reduce Sparsity?}
\label{sec:ablation-rubric-supervision}
Group-relative outcome advantages vanish when group rewards are identical. Under pure RL, Fig.~\ref{fig:opsd-beta}(c) compares progressive tasks with rubric rewards against endpoint QA with binary rewards. Coverage measures the fraction of four-trajectory groups with nonzero reward variance. The progressive task--reward design raises coverage from 36.5\% to 57.7\% and keeps it higher throughout training, because graded credit separates incomplete answers that binary correctness scores identically. MMSearch Image accuracy is higher at all six checkpoints, and the final gain across all five evaluation benchmarks is 2.1 percentage points. The gain in training signal thus exceeds the gain in accuracy.

\begin{table}[t]
    \centering
    \caption{\textbf{Experience-guided learning and prompting.} (a) Training-only guidance. (b) Frozen SFT with prompt guidance on a separate cohort. Quality scores are percentages; parentheses show changes from None (pp). Bold marks the highest quality or lowest resource use per panel.}
    \label{tab:pi-guidance}
    \label{tab:opsd-pi-source}
    \label{tab:pi-generation}
    \footnotesize
    \renewcommand{\arraystretch}{1.18}
    \setlength{\tabcolsep}{1.5pt}
    \begin{tabular*}{\linewidth}{@{}@{\extracolsep{\fill}}lcccccIcccc@{}}
        \toprule
        \tblrow{tablegroup}\textbf{Guidance} & \multicolumn{5}{c}{\textbf{(a) Training-time guidance}} & \multicolumn{4}{c}{\textbf{(b) Frozen SFT + prompt guidance}} \\
        \cmidrule(lr){2-6}\cmidrule(lr){7-10}
         & \textbf{AVG} & \textbf{MMSearch} & \textbf{BC-VL} & \textbf{MMSearch+} & \textbf{Turns} $\downarrow$ & \textbf{Accuracy} & \textbf{Rubric score} & \textbf{Turns} $\downarrow$ & \textbf{Output tokens} $\downarrow$ \\
        \cmidrule(lr){1-6}\cmidrule(lr){7-10}
        None & 59.8 & 78.9 & 56.0 & 31.0 & 6.0 & 55.7 & 66.5 & 5.8 & 912.4 \\
        \tblrow{tablefocus}Successful trajectory & \textbf{62.8}\tdelta{+3.0} & \textbf{79.5} & \textbf{62.0} & \textbf{35.0} & 6.3 & 80.1\tdelta{+24.4} & 89.5 & 3.5 & 589.3 \\
        Reference answer & 50.9\tdelta{-8.9} & 69.0 & 50.0 & 21.0 & \textbf{5.6} & \textbf{98.8}\tdelta{+43.1} & \textbf{99.4} & \textbf{3.2} & \textbf{499.4} \\
        \bottomrule
    \end{tabular*}
\end{table}

\subsubsection{What Experience Should the Self-Teacher See?}
\label{sec:ablation-experience-content}
Useful prompt context need not improve training feedback. In the frozen-SFT diagnostic (Table~\ref{tab:pi-generation}(b)), reference answers raise accuracy to 98.8\%, while successful trajectories raise rubric credit to 89.5\%. Both help when supplied during generation. In training, evaluated without experience (Table~\ref{tab:opsd-pi-source}(a)), they diverge: trajectories improve AVG by 3.0 percentage points over RL, whereas answers reduce it by 8.9 points. One explanation is that an answer mainly raises the likelihood of tokens that state the target, whereas a trace also shows which search decisions produced the evidence. We therefore use action--observation traces for calibration, although reference answers are the stronger prompt.

\begin{table}[t]
    \centering
    \caption{\textbf{Token selection and granularity for advantage correction.} Checkmarks denote corrected tokens; RL optimizes all policy tokens. AVG changes are versus Default (pp); bold marks highest AVG or lowest resource use. Evaluation uses no experience.}
    \label{tab:opsd-correction-design}
    \setlength{\arrayrulewidth}{0.3pt}
    \footnotesize
    \renewcommand{\arraystretch}{1.18}
    \setlength{\tabcolsep}{2pt}
    \begin{tabular*}{\linewidth}{@{}@{\extracolsep{\fill}}lccccIccccIcc@{}}
        \toprule
        \tblrow{tablegroup}\textbf{Variant} & \multicolumn{4}{c}{\textbf{Advantage correction}} & \multicolumn{4}{c}{\textbf{Accuracy (\%)}} & \multicolumn{2}{c}{\textbf{Resource use}} \\
        \cmidrule(lr){2-5}\cmidrule(lr){6-9}\cmidrule(lr){10-11}
         & Reasoning & Action & Answer & Granularity & \textbf{AVG} & MMSearch & BC-VL & MMSearch+ & \textbf{Turns} $\downarrow$ & \textbf{Tool cost} $\downarrow$ \\
        \midrule
        \tblrow{tablefocus}Default & -- & $\checkmark$ & $\checkmark$ & Token-level & \textbf{62.8} & 79.5 & 62.0 & 35.0 & 6.3 & 10.3 \\
        Action only & -- & $\checkmark$ & -- & Token-level & 56.1\tdelta{-6.7} & 74.9 & 53.0 & 27.0 & \textbf{5.7} & 9.9 \\
        All tokens & $\checkmark$ & $\checkmark$ & $\checkmark$ & Token-level & 58.2\tdelta{-4.6} & 76.0 & 54.0 & 32.0 & 5.8 & \textbf{9.6} \\
        \cmidrule(lr){1-11}
        Step average & -- & $\checkmark$ & $\checkmark$ & Step-level & 59.8\tdelta{-3.0} & 77.2 & 57.0 & 33.0 & 6.5 & 11.9 \\
        \bottomrule
    \end{tabular*}
\end{table}

\subsubsection{Where Should Calibration Apply?}
\label{sec:ablation-correction-design}
Table~\ref{tab:opsd-correction-design} varies the corrected tokens and the granularity of their corrections. Including answers alongside actions improves AVG by 6.7 percentage points over actions alone. Extending correction to reasoning lowers AVG by 4.6 points. Because PI contains actions and observations but omits reasoning, its evidence aligns most directly with action and answer decisions. Token-level calibration also exceeds step averaging by 3.0 points, so tokens within the same step benefit from separate corrections.

\subsubsection{How Strong Should Guidance Be?}
\label{sec:ablation-guidance-strength}
Fig.~\ref{fig:opsd-beta}(a,b) varies the maximum strength $\beta$. Both $\beta=0.05$ and $\beta=0.1$ improve over RL on all three benchmarks, with $\beta=0.1$ best on average. At $\beta=0.2$, the correction magnitude grows relative to the RL advantage (panel~b), while accuracy falls below RL. A larger correction is thus not better supervision, and we use a moderate strength by default for token-level calibration.

\begin{table}[t]
    \centering
    \caption{\textbf{Resource bonuses and reward-judging cost.} (a) Accuracy (\%) and resources; absolute differences are from $\lambda=0$, and Mean averages the three benchmark accuracies equally. (b) Estimated judge cost per 60-update run, excluding other training costs. Bold marks the best value per metric; green marks favorable changes and red unfavorable ones.}
    \label{tab:quality-cost-ablation}
    \footnotesize
    \renewcommand{\arraystretch}{1.18}
    \setlength{\tabcolsep}{2.2pt}
    \begin{tabular*}{\linewidth}{@{}@{\extracolsep{\fill}}lccccIccIlcc@{}}
        \toprule
        \tblband
        \multicolumn{7}{c}{\textbf{(a) Resource-bonus weight}} & \multicolumn{3}{c}{\textbf{(b) Rubric-judging strategy}} \\
        \cmidrule(lr){1-7}\cmidrule(lr){8-10}
        $\lambda$ & \textbf{Mean} & MMSearch & BC-VL & MMSearch+ & \textbf{Turns} $\downarrow$ & \textbf{Tool cost} $\downarrow$ & \textbf{Judge} & \textbf{Cost (\$)} $\downarrow$ & \textbf{Score (\%)} $\uparrow$ \\
        \cmidrule(lr){1-7}\cmidrule(lr){8-10}
        $0$ & \textbf{55.0} & \textbf{81.9} & \textbf{55.0} & 28.0 & 6.0 & 9.2 & Jev & \textbf{0.11} & 67.25 \\
        \tblrow{tablefocus}$0.2$ & 54.6\tdelta{-0.4} & 80.7 & 53.0 & 30.0 & 5.3\tsaving{-0.7} & 8.6\tsaving{-0.6} & Jev $\rightarrow$ Gemini & 2.27 & \textbf{68.12} \\
        $0.5$ & 54.3\tdelta{-0.7} & 77.8 & 54.0 & \textbf{31.0} & \textbf{4.5}\tsaving{-1.5} & \textbf{6.4}\tsaving{-2.8} & Gemini 3.5 Flash & 22.56 & 67.95 \\
        \bottomrule
    \end{tabular*}
\end{table}

\subsubsection{Can Rubric Judging Be Made Affordable?}
\label{sec:ablation-rubric-judging}
Rubric assessment is a recurring training expense. In Table~\ref{tab:quality-cost-ablation}(b), the Jev-to-Gemini cascade reduces estimated judging expenditure by 89.94\% (\$22.56 to \$2.27), with downstream scores of 68.12\% for the cascade and 67.95\% for full LLM judging. Jev alone is cheapest but scores lowest. Sending only uncertain cases to the stronger judge keeps its benefit at a small fraction of the full-run cost.

\subsubsection{Do Resource Bonuses Improve Accuracy?}
\label{sec:ablation-efficiency-shaping}
\begin{figure}[t]
  \centering
  \includegraphics[width=0.86\linewidth]{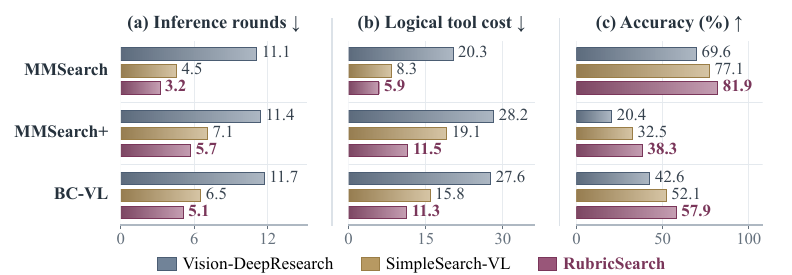}
  \caption{\textbf{Accuracy and interaction resources.} Mean inference rounds, logical tool operations, and accuracy on three search benchmarks. All methods are re-evaluated with our tools.}
  \label{fig:inference-resources}
\end{figure}

Table~\ref{tab:quality-cost-ablation}(a) varies the bonus weight $\lambda=\lambda_T=\lambda_C$. Turns and tool cost decrease monotonically. Raising $\lambda$ from 0.2 to 0.5 reduces turns by 14.7\% and tool cost by 25.9\%. Across the three increasing bonus settings, mean accuracy decreases from 55.0\% to 54.6\% and 54.3\%. Resource bonuses therefore control cost at a small price in accuracy, and both final models use $\lambda=0.2$. Re-evaluated in our tool environment on the same three search benchmarks (Fig.~\ref{fig:inference-resources}), RubricSearch achieves higher accuracy while using fewer inference rounds and logical tool operations than Vision-DeepResearch and SimpleSearch-VL. This resource reduction reflects the full training recipe, combining compact demonstrations, turn limits, and optional resource bonuses.

% !TeX root = ../main.tex
\FloatBarrier
\Needspace{8\baselineskip}
\section{Qualitative Analysis}
\label{sec:qualitative-analysis}
\label{sec:qualitative-visualization}

\begin{figure}[t]
    \centering
    \includegraphics[width=\textwidth]{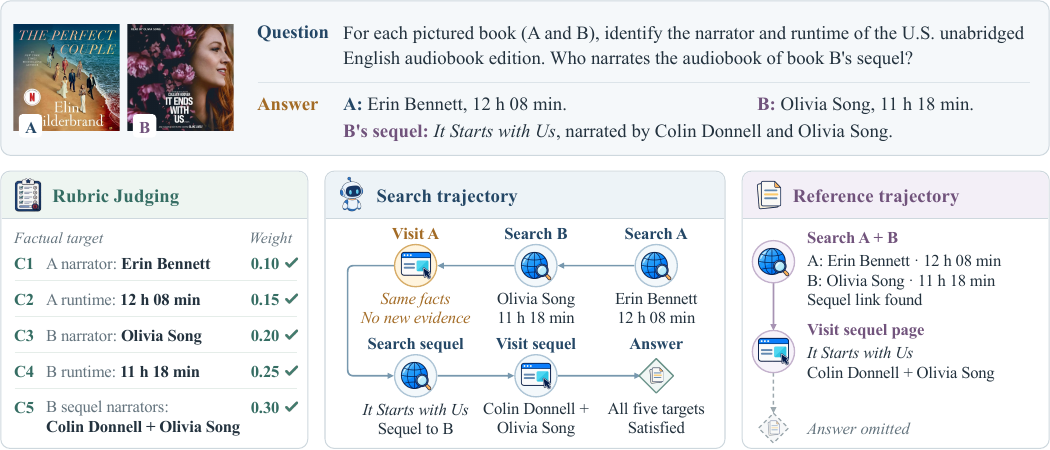}
    \caption{\textbf{Illustrative task and trajectories.} Circles mark tool calls, diamonds answer turns, arrows order, and checkmarks met criteria. The reference omits its answer turn and is used only for scoring.}
    \label{fig:training-example}
\end{figure}

Fig.~\ref{fig:training-example} illustrates the two signals on a reconstructed task. Five criteria assess narrators, runtimes, and a sequel. Both satisfy every criterion and receive the same rubric reward, although the five-call rollout repeats facts covered by the two-call reference. Rubric reward cannot distinguish them. Given the reference, the self-teacher can correct the redundant visit and answer differently despite their shared outcome advantage.

% !TeX root = ../main.tex
\FloatBarrier
\Needspace{8\baselineskip}
\section{Conclusion}
\label{sec:conclusion}
RubricSearch adds two kinds of feedback to outcome rewards for multimodal agentic search. Rubric rewards credit partial factual progress through criteria tied to facts explicitly requested by synthesized questions. Experience-conditioned self-distillation then calibrates the advantages of individual actions and answers with bounded corrections from a self-teacher; hierarchical trajectory distillation supplies compact demonstrations. Our analyses support graded task feedback, action--observation references, and moderate token-level corrections; resource bonuses control cost with a small accuracy trade-off. Inference requires neither auxiliary LLMs nor injected experience. The contrast between answers and traces suggests that privileged context helps credit assignment most when it records how evidence was obtained. Calibration needs prior same-task attempts; extending it to new tasks with experience retrieved from related tasks remains future work.

\bibliographystyle{ieeenat_fullname}
\bibliography{references}

% Appendix omitted from this release.
\end{document}